\pdfoutput=1

\documentclass[11pt]{article}

\usepackage[final]{coling}

\usepackage{times}
\usepackage{latexsym}
\usepackage{amsmath}
\usepackage{cleveref}
\usepackage{booktabs}
\usepackage{float}

\usepackage[T1]{fontenc}

\usepackage[utf8]{inputenc}

\usepackage{microtype}

\usepackage{inconsolata}

\usepackage{graphicx}

\title{Style-agnostic evaluation of ASR using multiple reference transcripts}

\author{Quinten McNamara, Miguel \'Angel del R\'io Fern\'andez, Nishchal Bhandari, \\
Martin Ratajczak, Danny Chen, Corey Miller \and Mig\"uel Jett\'e \\
Rev \\
\texttt{quinn@rev.com}}

\begin{document}
\maketitle
\begin{abstract}
Word error rate (WER) as a metric has a variety of limitations that have plagued the field of speech recognition. Evaluation datasets suffer from varying style, formality, and inherent ambiguity of the transcription task. In this work, we attempt to mitigate some of these differences by performing style-agnostic evaluation of ASR systems using multiple references transcribed under opposing style parameters. As a result, we find that existing WER reports are likely significantly over-estimating the number of contentful errors made by state-of-the-art ASR systems. In addition, we have found our multireference method to be a useful mechanism for comparing the quality of ASR models that differ in the stylistic makeup of their training data and target task.
\end{abstract}

\section{Introduction}
\label{sec:intro}
Word error rate (WER) has long been the primary metric for evaluating the performance of automatic speech recognition (ASR) systems. WER calculates the proportion of errors in a transcription, as measured by exact string equality against a ground truth reference. One common criticism of WER is the treatment of all errors as equally weighted, whether it is a simple stop word deletion or an egregious semantic error. To remedy this, much recent work in ASR evaluation has involved the development of assorted semantic-driven metrics \cite{rei2020comet, kim22p_interspeech}, mirroring earlier developments in machine translation (MT).

Another issue with WER is that human transcriptions of the same audio will often differ in ways that may not be of consequence to downstream users. The transcription choices made, be they light editing, adding/removing disfluencies, rephrasing, and so on, constitute what we call a transcript's \textbf{style} and can in many ways change the transcript without changing its semantic meaning. However, such differences can drastically affect WER, whether it is used by researchers aiming to improve their ASR system, or consumers seeking the most accurate ASR for their task.

\cite{heuser2024interspeech} created a corpus of multiple transcriptions of a subset of CORAAL \cite{kendall2018corpus} in order to explore stylistic differences in human and ASR transcriptions of the same audio. They identified differences in the way verbatimicity, morphosyntactic features, and reductions were represented in the various versions; these categories covered just under 30\% of all pairwise differences between transcriptions.

To help solve this problem of stylistic variation, we propose another parallel from MT research \cite{papineni-etal-2002-bleu, dreyer-marcu-2012-hyter} to alleviate treatment of these stylistic mismatches: multireferences. The term \textit{multiple reference(s)} (shortened as \textit{multireference}) is used to refer to multiple human transcriptions that serve as a reference against which machine transcriptions (or perhaps other human transcriptions) can be simultaneously evaluated.

The hypothesis underlying this concept is that in audio files of spontaneous speech of substantial length (e.g., greater than 30 seconds), such multiple references will not be identical. \cite{love2021specifying} provide a good overview of the various ways in which even linguistically-trained transcribers can exhibit great variation in transcription of the same audio. We expect the reference versions created in our work to reveal stylistic differences that affect calculation of WER.

\cite{ali2015} employ multireferences in the case of what they call a Non Standard Orthographic Language (NSO-L), Dialectal Arabic. NSO-Ls exhibit a great deal of orthographic variation which would cause WER, as normally calculated, to be higher than expected. However, by accepting as correct any of several multireferences, they are able to show reduced WERs, reflective of actual performance. We build upon this work by adopting a similar methodology for English ASR. As \cite{karita-etal-2023-lenient} point out, English orthographic variation is comparatively minor compared to Japanese. However, based on the findings of \cite{heuser2024interspeech}, we assert that the multireference approach is still necessary to disentangle stylistic variation in references.

Another motivation we would adduce in favor of multireferences is what we call ``sobriety''. For example, \cite{heuser2024interspeech} showed that Rev's accuracy is much better on Rev-produced transcripts than on those produced by Amberscript. In that work, Rev also appears superior to Whisper, but here we describe a \textit{GOLD WER} method, restricted to areas of cross-stylistic agreement, to provide a clearer picture of the relative accuracy between systems likely trained on different corpora and in different ways.

In this paper, we outline a finite-state transducer (FST) driven method of combining and scoring multireference transcripts that allows us to explore a new depth of style-agnostic accuracy in the world of speech recognition. In particular, we take two modern long-form datasets and produce human ground-truth transcripts that vary along some of the stylistic differences outlined in \cite{heuser2024interspeech}. We then benchmark a variety of state-of-the-art models against these new multireferences and present our interpretation of the impact on the field of ASR benchmarking.

\section{Additional Background}
\cite{rover1997}, known as ROVER, is an example of what we call \textit{multihypotheses}, working with multiple hypotheses from different ASR models to create alignment networks that reduce downstream  WER with a voting mechanism. Although our focus is on reference transcriptions, the word transition network (WTN) mechanism served as inspiration for the work presented here by illustrating a method for FST-driven solutions to the multiple alignment problem.

In the context of MT, \cite{dreyer-marcu-2012-hyter} indicate the oracular value of infinite translations: 
\begin{quote}
    If we had access to
the set of all correct translations of a given sentence,
we could measure the minimum distance between a
translation and the set. When a translation is perfect,
it can be found in the set, so it requires no editing to
produce a perfect translation.
\end{quote}

Restated in the context of ASR, if we knew all the possible ways people could (legitimately/plausibly/acceptably) transcribe an audio sample, then we could find a perfect machine transcription within that set. Multireferences move us toward the perfect oracle transcription set. The closer we get to including all possible transcriptions, the more accurate our scoring of machine transcriptions will be.

\cite{karita-etal-2023-lenient} employ a similar notion in their ``lenient'' CER proposal for Japanese, which employs multiple writing systems that can be used somewhat interchangeably for a subset of words. They asked humans to rate variants as ``acceptable'', ``great'' or ``wrong'', both in general and context-dependently. Importantly, they report that raters considered ``plausible'' 95.4\% of the variants their lenient scoring system admitted as correct. 

Any multireference scheme should attend to user acceptability of the variants considered correct for scoring. While \cite{ali2015} state that in English ``\textit{enough} is correct ... and \textit{enuf} is ... incorrect'', we believe some users, e.g. children, might prefer the informal spelling in some contexts (such as texting). We return to this point in \Cref{sec:methods}.

\cite{gandhi2022esb} acknowledge that there are many stylistic differences between benchmark datasets based on the formality and context of their creation. To address this, they suggest using a plethora of different benchmark datasets to get a complete picture, summarized by averaging across conditions. While we believe this to be a valid method when lacking additional resources, it does add a lot of noise to any error analysis efforts and masks the true accuracy of underlying systems with higher error rates.

\cite{faria22_interspeech} is a similar effort to show that WER on a standard English task, Switchboard, is actually lower than has been understood if one accounts for certain common alternate hypotheses. Their approach uses some of the various mechanisms that the \textit{sclite} tool provides for expressing alternates and synonyms. 

\section{Methods}
\label{sec:methods}
\begin{figure*}[ht]
    \centering
    \includegraphics[width=4cm]{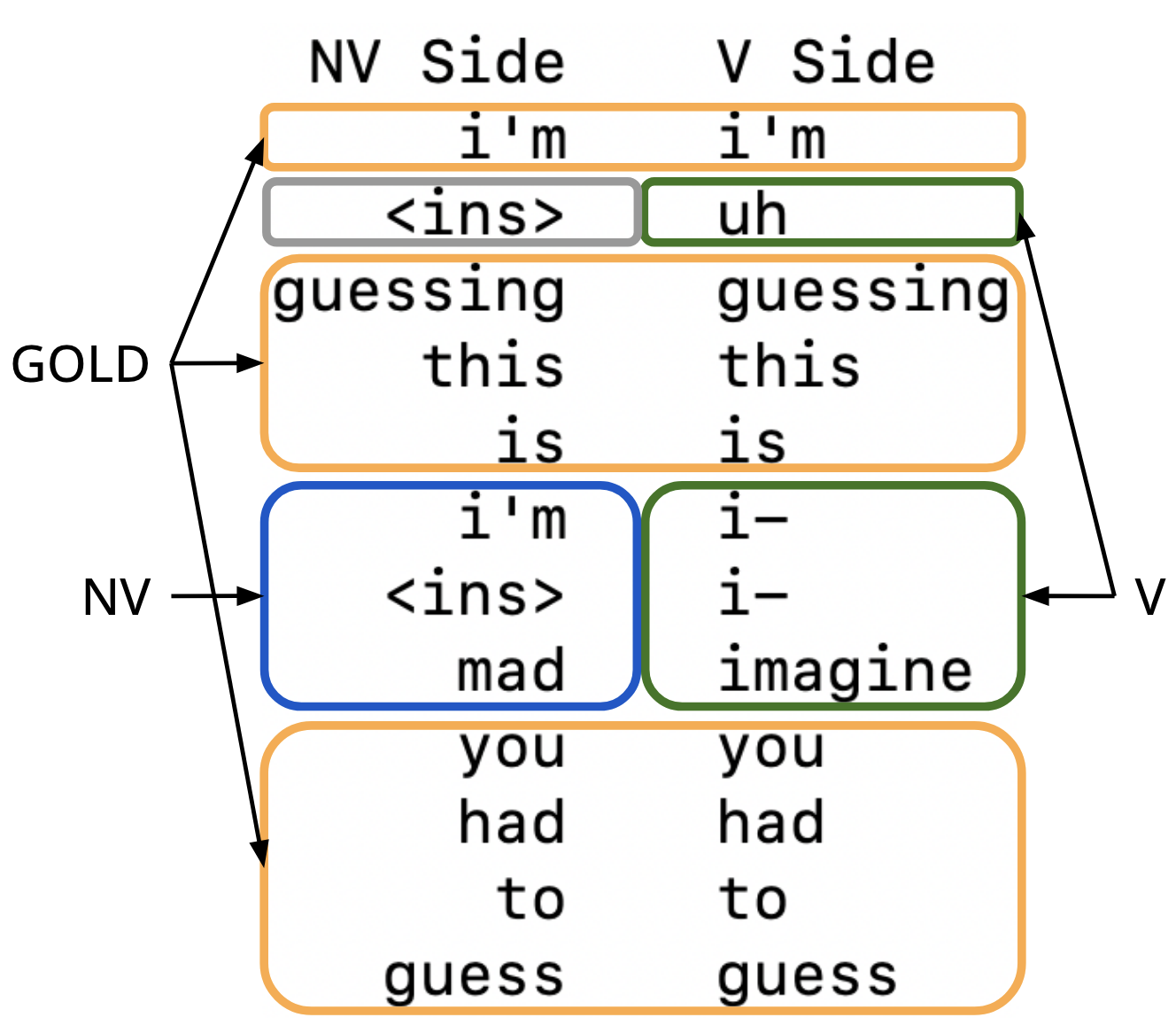}
    \includegraphics[width=\textwidth]{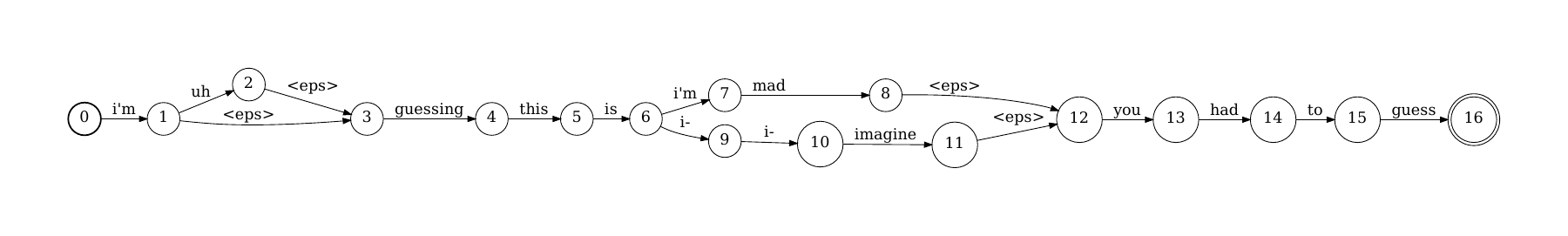}
    \caption{An alignment between two example transcripts and the resulting span-level multireference FST constructed. The three colors of boxes in the transcripts indicate the three tags discussed in \Cref{ssec:tagging}.}
    \label{fig:ex_multiref_fst}
\end{figure*}

\subsection{Datasets}
\label{ssec:datasets}
As part of this work, we provide multiple reference transcriptions for two distinct long-form evaluation sets: Rev16 and Earnings-22 Sub 10. Rev16, introduced by \cite{radford2023robust}  features 16 hours of podcast episodes composed entirely of American English speech. In contrast, Earnings-22 Sub 10, contains 10 hours of the Earnings-22 dataset \cite{earnings22} which consists of English language earnings calls from a variety of countries where English is a first, second, or other language. Thus the two datasets are contrasted in their orientation, entertainment vs. business, and accents, Amerian vs. World English.

We first gathered the original references for Rev16 from the blog post\footnote{\url{https://www.rev.com/blog/media-and-entertainment/podcast-transcription-benchmark-part-1}}. These references were said to be ``verbatim'', which according to the Rev.com style guide\footnote{\url{https://cf-public.rev.com/styleguide/transcription/Transcription+Style+Guide+v5.pdf}} includes ``exactly what you hear, including filler words, stutters, interjections (active listening) and repetitions.'' Secondly, we submitted the 16 podcast episodes again to Rev.com for human transcription, this time using default settings. This produced  ``lightly edit[ed] for readability'' nonverbatim transcriptions. From here on we use shorthand V for the verbatim copy of references and NV for the nonverbatim copy.

We selected Rev16 as one of our two datasets for a few reasons. Primarily, the error rates listed in \cite{radford2023robust} are still relatively high compared to other English ASR benchmarks such as Librispeech \cite{panayotov2015librispeech}, Switchboard \cite{fiscus20002000}, or TED-LIUM \cite{hernandez2018ted}. This could be due to the spontaneous nature, long-form structure, multi-speaker constitution, and presence of real-world modern slang and terminology. On the other hand, the audio is still of relatively high production quality with minimal background noise. This allows the subjectivity of transcription across stylistic differences to shine through as it limits the number of variations due to ambiguous transcription under difficult acoustic conditions.

The full Earnings-22 described in \cite{earnings22} was transcribed verbatim. We then took a random 10 hour subset to form Earnings-22 Sub 10 which contains those original verbatim transcriptions as well as an additional set of nonverbatim transcriptions, both sourced from Rev.com. In contrast to the high production quality of Rev16, Earnings-22 exhibits telephone quality, spontaneous speech, from a wide variety of native and nonnative accents of English. Thus it introduces the possibility of other factors clouding the ``cleaner'' picture potentially offered by Rev16.

\subsection{FST Method}
Our general method of using FSTs follows the fundamentals of \cite{dreyer-marcu-2012-hyter}. In contrast to their artificial collection of variants, we used natural collection via a word-level alignment of the two transcripts (V, NV) noted above. We used \textit{fstalign}\footnote{\url{https://github.com/revdotcom/fstalign/}} to produce such an alignment which is then constructed into a multireference FST\footnote{\url{https://github.com/revdotcom/fstalign/blob/develop/tools/sbs2fst.py}}, capturing the differences between transcriptions as shown in \Cref{fig:ex_multiref_fst}. We then run \textit{fstalign} again, passing in the multireference FST as the reference parameter and the machine transcription as the hypothesis.

\subsection{Tagging}
\label{ssec:tagging}
To better understand what multireference scoring reports, we implemented a mechanism to tag portions of the FST with information about its source reference (see \Cref{fig:ex_multiref_fst} for an example). The \textit{fstalign} tool is capable of reading these tags and aggregating the performance of each tag.

For the purpose of this paper, we utilized 3 tags:
\begin{itemize}
    \item \textbf{V}: This tag denotes the sections of a multireference that are \textit{only present} in the verbatim transcript.
    \item \textbf{NV}: This tag denotes the sections of a multireference that are \textit{only present} in the nonverbatim transcript.
    \item \textbf{GOLD}: This tag denotes the sections of a multireference that are in \textit{both} the verbatim and nonverbatim transcripts.
\end{itemize}
Models that tend to be more stylistically ``nonverbatim'' will likely align better to \textbf{NV} spans while the same is true for more ``verbatim'' models with \textbf{V} spans. The \textbf{GOLD} spans will be areas where both transcribers agreed and therefore have no bias toward any particular transcription style. In our results, we refer to the aggregate scores on these spans as the \textit{GOLD WER}, which allows us to more easily compare across models with many differences in style.

\subsection{Word-level vs span-level construction}
\label{section:word_span_level_construction}
Disagreements between references represent decision points where the multireference FST will fork. We took into consideration two different constructions of the FST object: word-level and span-level unions. The key distinction being that in word-level unions, each disagreeing word-pair results in a distinct FST branch, while span-level union will group consecutive disagreeing alignment pairs. Both constructions introduce the possibility of human-unverified paths in the multireference.

Empirically, we found that word-level unions achieved lower WER than span-level union WER. That being said, we also found word-level union allowed for many implausible hypotheses. For example, in \Cref{fig:ex_multiref_fst}, a word-level construction would consider the sentence ``i'm guessing this is i'm imagine you had to guess'' as a correct hypothesis despite it seeming highly implausible to utter. In the spirit of minimizing the number of accepted implausible hypotheses, we decided to proceed with span-level constructions and report only on their results. Analogous to \cite{karita-etal-2023-lenient}'s solicitation of plausibility judgments on orthographies, we could solicit grammaticality judgments on posited hypotheses. We leave this for future work.

\subsection{Bounds for span-level multireference WER}
\label{section:bounds_span-level_WER}
In this section, we give arguments for inequalities regarding multireference WER that can be derived from simple assumptions. 
First, by the definition of the minimum function, we know that the minimum of the WER over a set of references $R$ is lower than or equal to the WER for each individual reference $r$:
\begin{equation}
\min_{k \in R} W\!E\!R_k \leq W\!E\!R_r, \forall r \in R \label{eq:1}
\end{equation}

\cite{ali2015} reported empirical results that multireference WER is always lower than it is for a single reference and that it decreases with additional references. However, the construction in \cite{ali2015} differs from ours, as they have not used the FST method. The various paths encoded in a multireference FST can be converted into a set of references by exhaustively traversing all paths in the FST. In the span-level case, this new set of references, $S$, contains both the original human references and human-unverified references composed of combinations of spans from each human reference. The multireference WER (MWER) calculation algorithm should then pick amongst this set of references, selecting the one that represents the minimum WER path:
\begin{equation}
    M\!W\!E\!R_{spans} = \min_{j \in S} W\!E\!R_j \label{eq:2}
\end{equation}

Based on the definition of the minimum function on sets, we know that the minimum of the WER over a super-set of references can only be lower than or equal to that of the subset. The span-level FST encodes a super-set of references $S \supseteq R$ containing the human references $R$. Thus:
\begin{equation}
\min_{j \in S} W\!E\!R_j \leq \min_{k \in R} W\!E\!R_k. \label{eq:3}
\end{equation}

By the same logic, the word-level FST can contain even more human-unverified references and the reference list $W$ is a super-set of references $S$: $W \supseteq S$. Therefore, we can conclude that the word-level multireference WER is lower than or equal to the span-level one and serves as the lower bound.

In summary, we can provide WER inequalities:
\begin{equation}
    \begin{split}
        M\!W\!E\!R_{words} \leq M&\!W\!E\!R_{spans} \leq W\!E\!R_r
        \\
        \forall r &\in R
    \label{eq:5}
    \end{split}
\end{equation}
which can serve as lower and upper bounds for the span-level multireference WER. \Cref{tab:rev16results} empirically confirms the upper bounds for the span-level multireference WER. Apart from an anomaly exhibited by Canary-1B discussed in \Cref{sec:results}, these are also confirmed by \Cref{tab:Earnings-22results}.

\section{Results}
\label{sec:results}

\begin{table*}[!pt]
    \renewcommand\thetable{2a}
    \centering
    \textbf{Rev16} \\
    \begin{tabular}{c|c|c|c|c}
    \toprule
        Model & V-WER & NV-WER & Multireference WER & \textit{GOLD WER} \\
    \midrule
        Canary-1B & 13.19 & 13.87 & 9.00 & 7.90 \\
        Parakeet-CTC-1.1B & 12.02 & 12.90 & 7.25 & 6.10 \\
        Parakeet-TDT-1.1B & 12.90 & 13.91 & 9.08 & 8.00 \\
        OpenAI API & 11.21 & 9.62 & 5.45 & 4.43 \\
        RevAI API V & 8.03 & 12.33 & 5.43 & 4.35 \\
        RevAI API NV & 12.47 & 7.06 & 5.24 & 4.29 \\
    \end{tabular}
    \caption{Effect of multireference approach on WER benchmarking across open-source and commercial models on the Rev16 dataset. V-WER refers to the WER on the verbatim references only. Similarly for NV-WER.}
    \label{tab:rev16results}
\end{table*}

\begin{table*}[!pt]
    \renewcommand\thetable{2b}
    \centering
    \textbf{Earnings-22 Sub10} \\
    \begin{tabular}{c|c|c|c|c}
    \toprule
        Model & V-WER & NV-WER & Multireference WER & \textit{GOLD WER} \\
    \midrule
        Canary-1B$^{*}$ & 14.83 & 8.26 & 15.69 & 15.02 \\
        Parakeet-CTC-1.1B & 14.70 & 12.22 & 9.62 & 8.63 \\
        Parakeet-TDT-1.1B & 14.94 & 12.16 & 9.76 & 8.85 \\
        OpenAI API & 13.11 & 6.41 & 4.50 & 3.83 \\
        RevAI API V & 7.74 & 14.57 & 6.35 & 5.86 \\
        RevAI API NV & 14.06 & 5.37 & 5.29 & 4.62\\
    \end{tabular}
    \caption{Effect of multireference approach on WER benchmarking across open-source and commercial models on the Earnings-22 Sub10 dataset. V-WER refers to the WER on the verbatim references only. Similarly for NV-WER. \textit{*This Canary run had a significant hallucination in one file, impacting the results of this table.}}
    \label{tab:Earnings-22results}
\end{table*}

In \Cref{tab:ablation} we illustrate that standard rule-based practices of normalizing text reference and hypothesis transcripts are unsatisfactory in their ability to mitigate stylistic differences. For each condition, we applied the cleaning steps described to both the references and hypotheses in Rev16.

\begin{table}[H]
    \renewcommand\thetable{1}
    \small
    \centering
    \begin{tabular}{c|c|c|c}
    \toprule
        Condition & WER & INS & DEL \\
    \midrule
        Raw transcript (V) & 11.21 & 1.4 & 6.9 \\
        + filler word removal & 10.39 & 1.4 & 6.0 \\
        + normalized scoring & 10.31 & 1.6 & 6.0 \\
        + stutter and repetition removal & 9.56 & 1.9 & 4.8 \\
        + filler phrase removal & 9.23 & 1.9 & 4.4 \\
    \midrule
        Raw transcript (NV) & 9.62 & 4.6 & 2.7 \\
    \midrule
        Multireference & \textbf{5.45} & \textbf{1.4} & \textbf{2.5}
    \end{tabular}
    \caption{Measured WER results on the Rev16 dataset against the OpenAI API (run in March 2024) across different transcript post-processing conditions.}
    \label{tab:ablation}
\end{table}

\begin{itemize}
    \item \textbf{Filler word removal}: Removal of ``uh'', ``um'', and variants.
    \item \textbf{Normalized scoring}: Follows the Whisper repository recipe for English transcript normalization\footnote{\url{https://github.com/openai/whisper/tree/main/whisper/normalizers}}. It includes numeric standardization, contraction expansion, and British to American spelling normalization among other cleanup operations.
    \item \textbf{Stutter and repetition removal}: This condition uses regular expressions to remove disfluencies (e.g. ``w- wh- what'') and repeated terms (``who who are you'').
    \item \textbf{Filler phrase removal}: Removes additional phrases that are typically used as filler, such as ``you know'' ``like'' and ``so''.
\end{itemize}

Despite this collection of rule-based steps to clean up transcripts, none of them reaches the level of the multireference approach with verbatim and nonverbatim transcript versions. Part of this can be attributed to residual verbatim artifacts in the transcript, as is evidenced by the strong gap in deletion rate between the NV reference and the highest amount of cleaning (+ filler phrase removal) condition. Anecdotal investigation reveals categories of errors that were not addressed via rule-based cleaning: repetitions of 3+ word spans, precise transcription of crosstalk that the ASR usually skips over, etc. While it is likely possible to iron out all of these gaps on a case-by-case basis, as \cite{faria22_interspeech} postulate, common practice typically does not extend beyond the rules tested here.

In \Cref{tab:rev16results} and \Cref{tab:Earnings-22results} we highlight results using the multireference approach on top-performing models from the Open ASR Leaderboard\footnote{\url{https://huggingface.co/spaces/hf-audio/open_asr_leaderboard}} and top-performing commercial ASR APIs. We assume that the OpenAI API contains some derivative of the Whisper Large V3 model, thus we do not benchmark those results separately. The RevAI API allows users to select between ``verbatim'' or ``nonverbatim'' style transcription -- we opted to report on the results of both. Given that we used Rev.com for creating the reference transcripts, there is likely a small bias towards that model being the most accurate due to sharing the same pool of transcriptionists. The references are available publicly for others to run evaluation on their own models and APIs alike\footnote{\url{https://github.com/revdotcom/speech-datasets/tree/main/multireferences}}.

Seemingly contradicting our proof in \Cref{section:bounds_span-level_WER}, the Canary model's performance in \Cref{tab:Earnings-22results} shows a multireference WER \textit{worse} than its V-WER and NV-WER. Further investigation shows that this is due to a large hallucination in one file causing upwards of 80\% WER and significantly increasing the WER (without that file, the multireference WER is 6.71 and the GOLD WER is 5.92). We attribute this to the nature of \textit{fstalign} which doesn't guarantee a globally optimal alignment but rather uses heuristics and local optimal alignments to improve computation time. Despite this, we hold that the results of this study on our new technique are still valid and have significant implications.

We want to call attention to the performance of RevAI's V and NV results; V-WER is lowest on the output of the V model, while NV-WER is lowest on the output of the NV model. While intuitively this makes sense, it highlights the bias that style has on evaluation: if we had only gotten one style-version of the transcript, Rev's model would \textit{always} look better if the style matched. This underlines the need for an agnostic approach like the multireference method that avoids errors due to transcription guidelines. We clearly see that this multireference approach gives significantly lower absolute values for WER. We believe this provides evidence that most current evaluations of ASR systems are overstating the number of remaining errors. The decreasing effect is even stronger for the \textit{GOLD WER}, as described in \Cref{ssec:tagging}, which focuses on spans of human agreement.

It is important to note that direct comparison between models using multireference WER is not perfect, as the denominators across them may vary widely. This is because models can align with varying numbers of words from the multireference. For example, as illustrated in \Cref{tab:rev16denominators} and \Cref{tab:Earnings-22denominators}, the number of reference words for Parakeet-CTC is always higher compared to OpenAI API. This gap comes from the fact that Parakeet-CTC appears to attempt to be more verbatim with its output compared to OpenAI. This is another area where looking at the \textit{GOLD WER} is crucial, as it isolates all hypotheses to be scored against the portions of the transcript that both humans agreed upon regardless of style; we get a much smaller gap in the number of reference words between Parakeet-CTC and OpenAI API for both datasets, reducing the gap between references by 75\%. The remaining gap is likely due to discrepancies in normalization of contractions and numbers.

\begin{table*}[!pt]
    \renewcommand\thetable{3a}
    \centering
    \textbf{Number of Reference Words in Rev16} \\
    \begin{tabular}{c|c|c}
    \toprule
        Model & Multireference WER & \textit{GOLD WER} \\
    \midrule
        Parakeet-CTC & 184,918 & 170,410 \\
        OpenAI API & 182,711 & 169,979 \\
    \bottomrule
        Relative Difference & 1.2\% & 0.3\%
    \end{tabular}
    \caption{Variation in number of references used to calculate WER in select conditions for Rev16. Relative difference is included comparing Parakeet-CTC to OpenAI.}
    \label{tab:rev16denominators}
\end{table*}

\begin{table*}[!pt]
    \renewcommand\thetable{3b}
    \centering
    \textbf{Number of Reference Words in Earnings-22 Sub10} \\
    \begin{tabular}{c|c|c}
    \toprule
        Model & Multireference WER & \textit{GOLD WER} \\
    \midrule
        Parakeet-CTC & 102,187 & 95,374 \\
        OpenAI API & 99,790 & 94,814 \\
    \bottomrule
        Relative Difference & 2.4\% & 0.6\%
    \end{tabular}
    \caption{Variation in number of references used to calculate WER in select conditions for Earnings-22 Sub10. Relative difference is included comparing Parakeet-CTC to OpenAI.}
    \label{tab:Earnings-22denominators}
\end{table*}



The importance of \textit{GOLD WER} is echoed in the results between the OpenAI and RevAI APIs as seen in \Cref{tab:rev16results} and \Cref{tab:Earnings-22results}. The models underlying each of these APIs were likely trained on radically different datasets with large stylistic differences. As we see in the V-WER and NV-WER results, OpenAI is significantly more biased towards nonverbatim output while the RevAI is inversely biased towards verbatim output. However, when we boil down the comparison to \textit{GOLD WER}, the WER gap between the two becomes significantly smaller. Further demonstrating the value of these two metrics, \Cref{tab:Earnings-22results} shows something surprising. Despite RevAI having the lowest V-WER and NV-WER, OpenAI by far has the lowest multireference WER and \textit{GOLD WER}. We interpret this as follows: RevAI is best able to capture the verbatim and nonverbatim styles of the transcripts while OpenAI is best able to capture the transcript \textit{content}.

When developing models based on a consistent style of data, we believe that comparisons would best be done with the Multireference WER metric. This is because it would be important to identify style-specific errors and \textit{GOLD WER} may create a blind spot for practitioners. We leave this for future investigations into the technique.

\section{Conclusions}
\label{sec:conclusions}
Inspired by work from machine translation and motivated by recent developments in benchmarking speech recognition across transcription styles, we introduce a simple method and example datasets for isolating style-agnostic errors with multiple human references. We encourage others to build upon this work by further evaluating on these datasets and growing our understanding of inter-transcriber variation. We acknowledge that our methods are likely hard to adopt in the speech recognition field because sourcing multireference transcripts would likely be twice as expensive as sourcing single references. However, we hope the discrepancies illustrated in our results comparing traditional benchmarking with multireference benchmarking serve as sufficient motivation to push adoption forward. Particularly as error rates continue to decline and we approach human levels of machine transcription, we maintain that improved benchmarks are increasingly necessary.

Our approach is not without limitations. As we note in \Cref{section:bounds_span-level_WER}, we know the upper bound of multireference WER will get smaller and smaller if we were to add more references. In our current setup, the flexibility of the algorithm to select any of the human-produced paths in the reference creates a double-edged sword of masking away human error. We believe that extending our approach into efforts with 3+ references should include a voting or weighting scheme that ultimately has the objective of \textit{correcting} human errors with additional references, instead of masking them out.

\bibliography{custom}




\end{document}